%% file: acl19slate.tex
\documentclass[11pt,a4paper]{article}
\usepackage[hyperref]{acl19slate}
\usepackage{times}
\usepackage{latexsym}

\usepackage{url}

\usepackage{amsmath}
\usepackage{booktabs}
\usepackage{multirow}
\usepackage{xspace}
\usepackage{graphicx}
\usepackage{color}
\usepackage{enumitem}
\usepackage{xcolor,colortbl}
\usepackage{multicol}
\usepackage{subcaption}

\hypersetup{pdftitle={slate: A Super-Lightweight Annotation Tool for Experts},
pdfauthor={Jonathan K. Kummerfeld},
pdfsubject={Natural Language Processing},
pdfpagelayout=OneColumn}

\newcommand{\myeg}{e.g.\@\xspace}

\newcommand{\tightparagraph}[1]{\noindent\textbf{#1}}

\newcommand{\unixcmd}[1]{\texttt{#1}}

\aclfinalcopy 
\def\aclpaperid{12} 

\newcommand\Slate{\textsc{slate}\xspace}
\newcommand\slate{\textsc{slate}\xspace}

\title{\slate: A Super-Lightweight Annotation Tool for Experts}

\author{
  Jonathan K. Kummerfeld \\
  Computer Science \& Engineering \\
  University of Michigan \\
  Ann Arbor, MI, USA \\
  {\tt jkummerf@umich.edu}
}

\date{}

\begin{document}
\maketitle

\begin{abstract}

  Many annotation tools have been developed, covering a wide variety of tasks and providing features like user management, pre-processing, and automatic labeling.
  However, all of these tools use Graphical User Interfaces, and often require substantial effort to install and configure.
  This paper presents a new annotation tool that is designed to fill the niche of a lightweight interface for users with a terminal-based workflow.
  \Slate supports annotation at different scales (spans of characters, tokens, and lines, or a document) and of different types (free text, labels, and links), with easily customisable keybindings, and unicode support.
  In a user study comparing with other tools it was consistently the easiest to install and use.
  \Slate fills a need not met by existing systems, and has already been used to annotate two corpora, one of which involved over 250 hours of annotation effort.

\end{abstract}

\section{Introduction}

Specialised text annotation software improves efficiency and consistency by constraining user actions and providing an effective interface.
While current annotation tools vary in the types of annotation supported and other features, they are all built with direct manipulation via a Graphical User Interface (GUI).
This approach has the advantage that it is easy for users who are not computer experts, but also shapes the design of tools to become large, complex pieces of software that are time-consuming to set up and difficult to modify.

We present a lightweight alternative that is not intended to cover all use-cases, but rather fills a specific niche: annotation in a terminal-based workflow.
This goal guided the design to differ from prior systems in several ways.
First, we use a text-based interface that uses almost the entire screen to display documents.
This focuses attention on the data and means the interface can easily scale to assist vision-impaired annotators.
Second, we minimise the time cost of installation by implementing the entire system in Python using built-in libraries.
Third, we follow the Unix Tools Philosophy \citep{unix} to write programs that do one thing well, with flat text formats.
In our case,
(1) the tool only does annotation, not tokenisation, automatic labeling, file management, etc, which are covered by other tools, and
(2) data is stored in a format that both people and command line tools like \texttt{grep} can easily read.

\Slate supports annotation of items that are continuous spans of either characters, tokens, lines, or documents.
For all item types there are three forms of annotation: labeling items with categories, writing free-text labels, or linking pairs of items.
Category labels are easily customisable, with no limit on the total number and the option to display a legend for reference.
All keybindings are customisable, and additional commands can be defined with relatively little code.
There is also an adjudication mode in which disagreements are displayed and resolved.

To compare with other tools we conducted a user study in which participants installed tools and completed a verb tagging task in a 623 word document.
When using \slate, participants finished the task in 13 minutes on average, with more than half spending 3 minutes or less on setup.

Two research projects have used \slate for annotation:
token-level classification of 25,624 tokens of cybercriminal web forum data \citep{www17forums},
and line-level linking of 77,563 messages of chat data \citep{acl19disentangle}.

\begin{figure*}
  \begin{minipage}[b]{.48\linewidth}
  \centering
    \includegraphics[width=\linewidth]{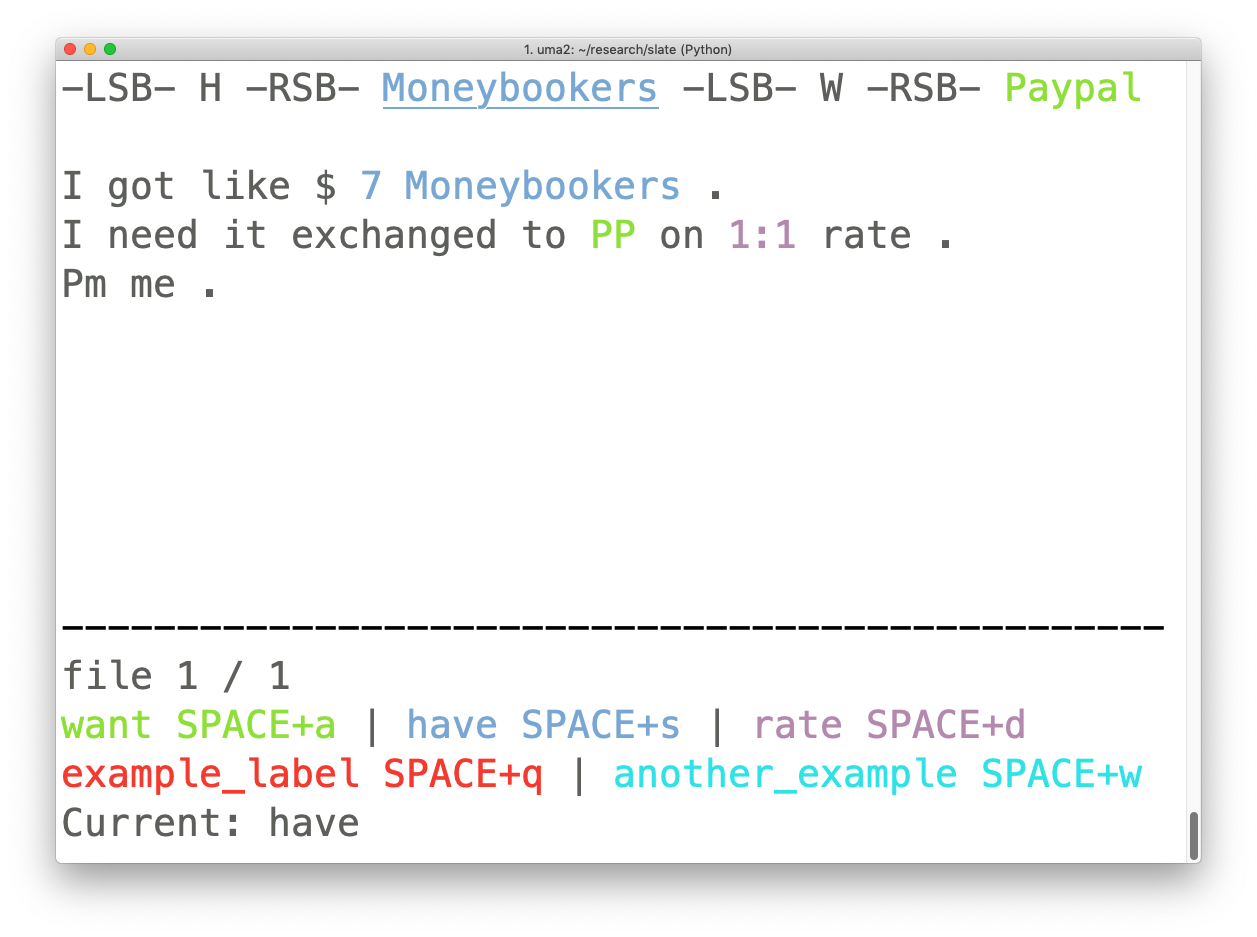}
  \subcaption{
    Annotation of a cybercriminal forum post.
    The underlined token is the one currently selected.
    The mapping from colours to labels and the keys to apply them are indicated by the legend at the bottom.
    Two extra labels were added for this picture, to show how the legend will wrap as needed.
    The information about progress, the legend, and the current item can be hidden if desired.
  }\label{fig:classify}
  \end{minipage}\hfill%
  \begin{minipage}[b]{.48\linewidth}
  \centering
    \includegraphics[width=\linewidth]{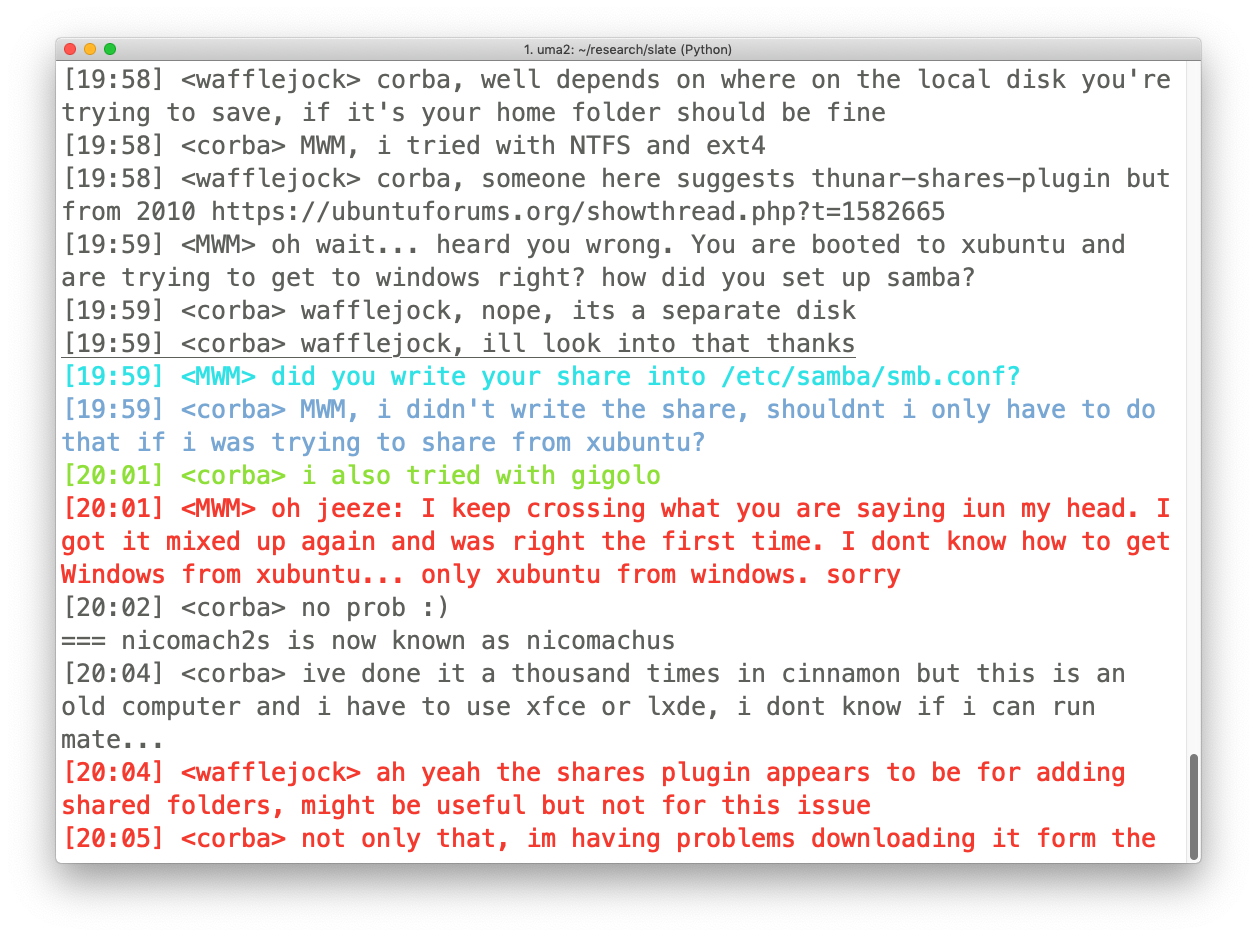}
  \subcaption{
    Adjudication of a linking task annotating reply-to relations on IRC.
    The current message in green, an antecedent that is agreed on is dark blue, one that is not agreed on is light blue, other messages with disagreements are red, and the underline indicates a message that is being considered for linking to the green message (incorrectly in this case). \\
  }\label{fig:adjudicate}
  \end{minipage}
  \caption{
  Screenshots of terminal windows with \slate running.
  We have intentionally used two different font sizes to show how the user interface can scale, which is helpful for visually impaired users.
  }\label{fig:screenshot}
\end{figure*}

\section{System Description}

Rather than specifying a task, such as named entity recognition, \slate is designed to flexibly support any annotation that can be formulated as one of three annotation types: applying categorical labels, writing free text, or linking portions of text.
These can be applied to items that are single documents, lines, tokens, or characters, and continuous spans of lines, tokens, and characters.
For example, from this perspective, NER is a task with categorical labels over continuous token spans.

The tool also supports adjudication of annotation disagreements.
Multiple sets of annotations are read in and compared to determine disagreements to be displayed to the user, which can then be resolved, producing a new annotation file.

In the process of describing the tool, we will refer to two datasets that it was used to annotate:\footnote{
No formal user satisfaction surveys were conducted, but the tool received positive feedback on both projects.
}

(1) \citet{www17forums} studied cybercriminal web forums.
Experts in computer security and NLP collaborated to label posts in which the user is trying to exchange money from one form into another.
For each post, we labeled tokens that expressed what was being offered, what was being requested, and the rate.
Three annotators labeled a set of 600 posts containing 25,624 tokens.
200 of the posts were triple annotated and disagreements were adjudicated using the tool.

(2) \citet{acl19disentangle} developed a new conversation disentanglement dataset an order of magnitude larger than all previously released datasets combined.
77,563 messages were annotated with links indicating the message(s) they were a response to.
For 11,100 messages, multiple annotations were collected and adjudicated using the tool.
Annotations took 7 to 11 seconds per message depending on the complexity of the discussion, and adjudication took 5 seconds per message (lower because many messages did not have a disagreement, but not extremely low because those that did were the difficult cases).
Overall, annotators spent approximately 240 hours on annotation and 15 hours on adjudication.

\input{comparison}

\tightparagraph{Display}
The interface is text-based and contained entirely within a terminal window.
By default, the entire terminal area displays the text being annotated, as shown in Figure~\ref{fig:adjudicate}.
There are also options to use space to display information about the current annotations, as shown in Figure~\ref{fig:classify}.

Colour is used to indicate annotations.
For categorical labels a different colour is used for each label, and a special colour is used to indicate when multiple labels have been assigned to the same item.
For linking, one colour is used to indicate the item being linked, another is used to indicate what items are currently linked to it, and a third is used to indicate any item that has a link.

The most frequently needed visual customisation is changing the colours used to indicate labels when assigning categories.
These colours are specified in a simple text file.
More substantial changes, such as a different colour to indicate disagreements, requires changing the code, but colours for each situation are defined in one place with intuitive names.

\tightparagraph{Input}
All interaction occurs via the keyboard.
Commands are designed to be intuitive, \myeg arrow keys change which item is selected.
Keybindings can be modified to suit user preferences and multi-key commands can be defined, like the labels in Figure~\ref{fig:classify} (\myeg \texttt{SPACE+a}), providing flexibility and an unlimited set of combinations.

Basic commands cover selecting a span, assigning an annotation, and undoing annotations.
Exact string search is included, which was used in the disentanglement project to search for previous messages by a given user.
Typing a number before a command will apply the command that many times.
A range of commands exist to toggle properties of the view, including line numbering, the legend, showing the label for the selected item, and progress through a set of files.

\tightparagraph{Data}
Items in annotations are represented by tuples of integers and labels are represented as strings.
Tokens are defined by splitting on whitespace, and characters are counted with separate numbering for each token.
This makes the annotations easy to read in and interpret.

Externally, annotations are saved in stand-off files, with one annotation per line.
This makes them easy to process with command line tools, \myeg using \unixcmd{wc} to count the number of annotations.

\tightparagraph{Extension}
The system is written with a modular design intended to be easily modifiable.
For example, at one point in the disentanglement project we wanted to go back and check messages that started conversations.
This involved adding 21 lines of code, extending the existing search commands to jump to the next line that started a conversation.

\tightparagraph{Internal Architecture}
The system is written in 2,300 lines of Python, with extensive use of the standard built-in curses library for input and display handling.
Setting up the tool involves downloading and unzipping the code.
The tool can then be run from anywhere, does not require administrator privileges, and will only make modifications to the local directory it is being run from.

\tightparagraph{Dependencies}
On macOS and all Linux distributions we are aware of, there is nothing else to install: Python and the relevant libraries are already installed.
On Windows, Python (either 2 or 3) and the curses library need to be installed.

\section{Related Work}

Table~\ref{tab:comparison} presents a comparison of a range of annotation tools.\footnote{
  Non-academic tools also focus on GUIs, including open source tools
  (\myeg \href{https://github.com/chakki-works/doccano}{DocAnno}, \href{https://github.com/proycon/flat}{Flat}), and  
  commercial tools
  (\myeg \href{https://www.tagtog.net/}{TagTog}, \href{https://prodi.gy/}{Prodigy}, \href{http://mat-annotation.sourceforge.net/}{MAT}, \href{https://www.lighttag.io/}{LightTag}, \href{https://dataturks.com/}{DataTurks}).
}
Note that all prior work has focused on graphical user interfaces.
The top section shows the tools we consider in our user study, chosen because they are widely used (brat and GATE) or have a similar design motivation (YEDDA).

\tightparagraph{brat} \citep{brat}, is a synchronous web-based system with a central server that users connect to using a web browser.
The tool supports annotation of spans of text and relations between them, either binary relations, equivalence classes, or n-ary associations.
It provides a search mechanism, automatic validation, tagging with external tools, multi-lingual processing, and a comparison mode that places two annotations side-by-side.
The tool has been used to annotate a range of datasets, particularly for information extraction from biomedical documents.

\tightparagraph{GATE} \citep{gate}, is both a desktop application and a web-based system.
It is designed to support the entire lifecycle of a project, including data preparation, schema creation, annotation, adjudication, data storage and use.
To achieve this, it contains a wide array of components, covering various annotation types and tools to define workflows that determine the stages of annotation of a document.
The system has been used in a range of projects in both academic and commercial settings, including with users who did not have a computer science background.

\tightparagraph{YEDDA} \citep{yedda}, is a desktop application that supports annotation of character spans with up to 8 categories using a GUI.
It is designed to be lightweight, with no external dependencies, though it only supports Python 2.
It has a built-in recommendation system that automatically proposes labels based on annotations so far, which \citet{yedda} found decreased named entity recognition annotation time by 16\%.
For input, the tool supports both selection with the mouse and specifying a sequence of character-level annotations with the keyboard, \myeg \texttt{2c3b} to assign the label \texttt{c} to the next two characters and \texttt{b} to the three after that.
The tool also has an analysis component that produces a LaTeX document comparing a pair of annotations of a document, or F-scores for all pairs of annotations.
It does not support annotation of links or adjudication of disagreements.

\section{User Study}

We conducted a user study\footnote{
  Approved by the Michigan IRB, ID: HUM00155689.
} to investigate the effort required to install and use four tools.
For each tool, participants had 25 minutes to install the tool and use it to identify verbs in a 623 word news article.\footnote{
  We also considered having participants do a linking task, specifying the antecedents of pronouns, but found there was not enough time (also, linking is not supported by YEDDA).
}
We measured how long they took to install the tool and to finish the task.
After each tool, participants completed a survey asking about their experience installing and using the tool.

We set up the study to simulate a real usage scenario as closely as possible.
Participants were computer science graduate students and all but one used their own computer.
Four participants used macOS and four used Ubuntu.
Every participant used all four tools.
To reduce bias due to fatigue (which could make later tools appear slower) and familiarity with the news article (which could make later tools appear faster), the order of tools varied so that each one occurred 1st, 2nd, 3rd, and 4th once for each operating system.

\begin{table}
  \centering
  \small
  \begin{tabular}{lrrr}
    \toprule
           & \multicolumn{2}{c}{Time (minutes)} \\
    Tool   & Ubuntu & macOS \\
    \midrule
    \slate & 10 & 16 \\
    YEDDA  & 14 & 14 \\
    GATE   & 21 & 22 \\
    brat   &  - &  - \\
    \bottomrule
  \end{tabular}
  \caption{\label{tab:study}
  Average time for users to set up the tool and identify verbs in a 623 word news article.
  Only one participant managed to install and use brat, taking 18 minutes on Ubuntu.
  The differences between GATE and either \slate or YEDDA are significant at the 0.01 level according to a t-test.
  }
\end{table}

Table~\ref{tab:study} presents the time required to install each tool and complete the first annotation task.
We combined these two because some participants read usage instructions while installing the tool, while others went back and forth during initial annotations.
\slate and YEDDA are comparable in effort, which fits with their common design as simple tools with minimal dependencies.
Participants had great difficulty with brat, with only two managing to install it, one just as their time finished.

We also had an additional participant try using the tools in Windows.
They were able to install and use \slate and GATE, but did not complete the task on either.
GATE had the easiest set up process, with a provided installer.
\slate required the curses package to be installed.
They experienced difficulty with YEDDA because it is not Python 3 compatible, and rewriting the code to make it compatible led to other issues.
brat only supports Windows via the use of a virtual machine.

\subsection{Feedback Summary}

\paragraph{brat}
Participants got stuck on various issues during installation and generally commented that they wanted more information in the documentation.
The two most common issues were being unable to get the system to communicate with apache, and being unable to load files to annotate.

\paragraph{GATE}
Participants often had to try more than one installation method.
Six participants commented on lag or missed clicks during annotation, and six commented that the documentation was not helpful for the process of installing the tool and understanding the user interface.

\paragraph{YEDDA}
All participants found installation easy, though three commented that it required familiarity with git, and one recommended adding it to the python package index.
Every participant commented that the recommendation mode produced many false positives, making it inconvenient.
Four commented that the way labels appear during annotation shifted the text in a way that made them lose their place or made it hard to read.
All Ubuntu participants commented that the system scrolled to the top of the document after each annotation.

\paragraph{\slate}
All participants found installation easy, though one recommended adding it to the python package index.
One participant commented that they wished they could use their cursor as well, one suggested adding an explicit indication that saving was successful, one found the selected span hard to see, and one was surprised that pressing the down arrow selected the same position on the next line in terms of tokens, rather than visually.

\paragraph{}
Each of these tools has strengths and weaknesses, and this study was designed to test the specific gap our tool was designed to address.
For projects with annotators who are not computer experts, we expect GUIs, and in particular web interfaces like the one provided by brat, will remain dominant.
However, the results demonstrate that \slate effectively fills the gap of an easy to use and efficient terminal-based tool.

\section{Future Development}

There are many directions in which we hope to develop \slate further, by collaborating with other researchers and the open-source community.
The system is designed to be simple to modify and extend, with some possible next steps including:

\tightparagraph{Multi-lingual Support}
The tool can already display a range of scripts given Python's unicode support, but to properly support annotation would require modifications.
For example, changing the direction of script writing would involve adjusting the visual display module, while adding support for selecting parts of glyphs would require modifying the data representation.

\tightparagraph{Accessibility}
For users who require large fonts (or prefer small ones) the tool is extremely flexible and convenient.
However, the extensive use of colour could be a problem for colour-blind users.
The colours can be modified by editing a simple configuration file, or colour could be avoided entirely by modifying the display code to display inline labels.

\tightparagraph{Data Formats}
Support for additional data formats, such as CoNLL-U, and \citet{data-iso}.

\tightparagraph{Mouse Input}
While the motivation for this tool was to make a text-based interface controlled by the keyboard, the curses library does process mouse inputs, making it possible to add support for the mouse in the future.

\section{Conclusion}

A wide range of effective annotation tools already exist, but they all use a GUI and many involve substantial effort to set up.
\slate is designed for terminal-users who want a fast, easy to install, and flexible annotation tool.
It supports a range of annotation types, and adjudication of disagreements in annotations.
The code is publicly available\footnote{\url{http://jkk.name/slate/}} under a permissive open-source license.
The tool has already been used in two research projects, including one that involved over 250 hours of annotation.

\section*{Acknowledgements}

Thank you to Will Radford and the anonymous reviewers for helpful suggestions.
Also thanks to the study participants, and to the annotators who have used the tool.
This material is based in part on work supported by IBM as part of the Sapphire Project at the University of Michigan, and by ONR under MURI grant N000140911081.
Any opinions, findings, conclusions or recommendations expressed above are those of the author and do not necessarily reflect the views of the sponsors.

\bibliography{acl19slate}
\bibliographystyle{acl19slate}

\end{document}

%% file: comparison.tex
\begin{table*}
  \centering
  \small
  \begin{tabular}{lccccc}
    \toprule
           & Annotation &  Adjud- & External & Programming &      User \\
    System &      Types & ication & Dependencies & Language & Interface \\
    \midrule
    \slate & Classify, Link & Yes & - & Python & Terminal \\
    brat \citep{brat} & Classify, Link & Yes & apache & Python, Javascript & GUI \\
    GATE \citep{gate} & Classify, Link & Yes & - & Java & GUI \\
    YEDDA	\citep{yedda} & Classify &  - & - & Python & GUI \\
    \midrule
    ANALEC \citep{analec} & Classify, Link & - & - & Java & GUI \\
    Anafora \citep{anafora} & Classify, Link & Yes & - & Python & GUI \\
    CAT \citep{cat} & Classify & - & apache & Java & GUI \\
    Chooser \citep{chooser} & Classify, Link & - & - & C++, Python, Perl & GUI \\
    CorA \citep{cora} & Classify & - & - & PHP, JavaScript & GUI \\
    Djangology \citep{djangology} & Classify & Yes & Django & Python & GUI \\
    eHost \citep{ehost} & Classify, Link & Yes & - & Java & GUI \\
    Glozz \citep{glozz} & Classify, Link & Yes & - & Java & GUI \\
    GraphAnno \citep{graphanno} & Classify, Link & - & - & Ruby & GUI \\
    Inforex \citep{inforex} & Classify, Link & - & - & JavaScript & GUI \\
    Knowtator	\citep{knowtator} & Classify, Link & Yes & Prot\'eg\'e & Java & GUI \\
    MAE and MAI \citep{mae-mai} & Classify, Link & Yes & - & Java & GUI \\
    MMAX2 \citep{mmax} & Classify, Link & - & - & Java & GUI \\
    PALinkA \citep{palinka} & Classify, Link & - & - & Java & GUI \\
    SAWT \citep{sawt} & Classify & - & - & Python, PHP & GUI \\
    SYNC3 \citep{sync3} & Classify & Yes & Ellogon & C & GUI \\
    Stanford \citep{stanford} & Classify & - & - & Java & GUI \\
    UAM \citep{uam} & Classify & Yes & - & Java & GUI \\
    WAT-SL \citep{wat-sl} & Classify & Yes & apache & Java & GUI \\
    WebAnno \citep{webanno} & Classify, Link & Yes & - & Java & GUI \\
    WordFreak	\citep{wordfreak} & Classify, Link & Yes & - & Java & GUI \\
    \bottomrule
  \end{tabular}
  \caption{\label{tab:comparison}
  A comparison of annotation tools in terms of properties of interest (in some cases tools support extra types of annotation, \myeg syntax, that we do not consider here).
  }
\end{table*}